\documentclass[journal]{IEEEtran}
\usepackage{cite}
\usepackage{graphicx}
\usepackage{latexsym}
\usepackage{float}
\usepackage{hyperref}
\usepackage{array}
\usepackage{amsmath,amssymb,amsfonts}
\usepackage{booktabs}
\usepackage{multirow}
\usepackage{tabularx}
\usepackage{xcolor}
\usepackage[linesnumbered,ruled]{algorithm2e}
\usepackage{algpseudocode}

\begin{document}
\title{Facial Anatomical Landmark Detection using Regularized Transfer Learning with Application to Fetal Alcohol Syndrome Recognition
}

\author{Zeyu~Fu,~\IEEEmembership{Member,~IEEE,}
       Jianbo~Jiao,~\IEEEmembership{Member,~IEEE,}
       Michael~Suttie,
       and J. Alison~Noble.
\thanks{This work was done in conjunction with the Collaborative Initiative on Fetal Alcohol Spectrum Disorders (CIFASD), which is funded by grants from the National Institute on Alcohol Abuse and Alcoholism (NIAAA).
	This work was supported by NIH grant U01AA014809 and EPSRC grant EP/M013774/1.
}
\thanks{Ethical approval for this study was obtained from Oxford Tropical Research Ethics Committee, University of Oxford (Date of approval: 31/03/17 and Protocol number: 519-17).}
\thanks{Z. Fu, J. Jiao and J.A. Noble are with the Department of Engineering Science, University of Oxford, Oxford, UK, Emails: \{zeyu.fu, jianbo.jiao, alison.noble\}@eng.ox.ac.uk.}
\thanks{M. Suttie is with the Nuffield Department of Women's and Reproductive Health, Big Data Institute, University of Oxford, Oxford, UK,  E-mail: michael.suttie@wrh.ox.ac.uk.}
}

\maketitle
\begin{abstract}
Fetal alcohol syndrome (FAS) caused by prenatal alcohol exposure can result in a series of cranio-facial anomalies, and behavioral and neurocognitive problems. Current diagnosis of FAS is typically done by identifying a set of facial characteristics, which are often obtained by manual examination. Anatomical landmark detection, which provides rich geometric information, is important to detect the presence of FAS associated facial anomalies.
This imaging application is characterized by large variations in data appearance and limited availability of labeled data. 
Current deep learning-based heatmap regression methods designed for facial landmark detection in natural images assume availability of large datasets and are therefore not well-suited for this application. To address this restriction, we develop a new regularized transfer learning approach that exploits the knowledge of a network learned on large facial recognition datasets.
In contrast to standard transfer learning which focuses on adjusting the pre-trained weights, the proposed learning approach regularizes the model behavior. It explicitly reuses the rich visual semantics of a domain-similar source model on the target task data as an additional supervisory signal for regularizing landmark detection optimization.
Specifically, we develop four regularization constraints for the proposed transfer learning, including constraining the feature outputs from classification and intermediate layers, as well as matching activation attention maps in both spatial and channel levels.
Experimental evaluation on a collected clinical imaging dataset demonstrate that the proposed approach can effectively improve model generalizability under limited training samples, and is advantageous to other approaches in the literature.
\end{abstract}

\IEEEpeerreviewmaketitle
\section{Introduction}
\label{Introduction}
Accurate localization of anatomical landmarks is a fundamental step for numerous medical imaging applications such as image registration and shape analysis \cite{regression-forest,Deep-Anatomical-Cephalometric,regression-voting,Two-StageTask-Oriented,deep-landmark-medical}.
It also has the potential to advance the early recognition of those affected by prenatal alcohol exposure \cite{FAS}, \cite{DSNT}.
Prenatal alcohol exposure (PAE) damages the developing central nervous system, resulting in a continuum of effects \cite{PAE}. Conditions that result from PAE are collectively known as fetal alcohol spectrum disorders (FASDs). Fetal alcohol syndrome (FAS) is the most severe outcome in this spectrum, characterized by restricted growth, neurobehavioral deficits, and facial dysmorphism \cite{FASD}.
The diagnosis of FAS requires at least 2 of 3 cardinal facial features; a smooth philtrum, a thin upper lip, and a reduced palpebral fissure length (PFL)\cite{PAE,FASD}.

\begin{figure}[t]
\centering
\setlength{\belowcaptionskip}{-0.5cm}
\includegraphics[width=0.85\linewidth]{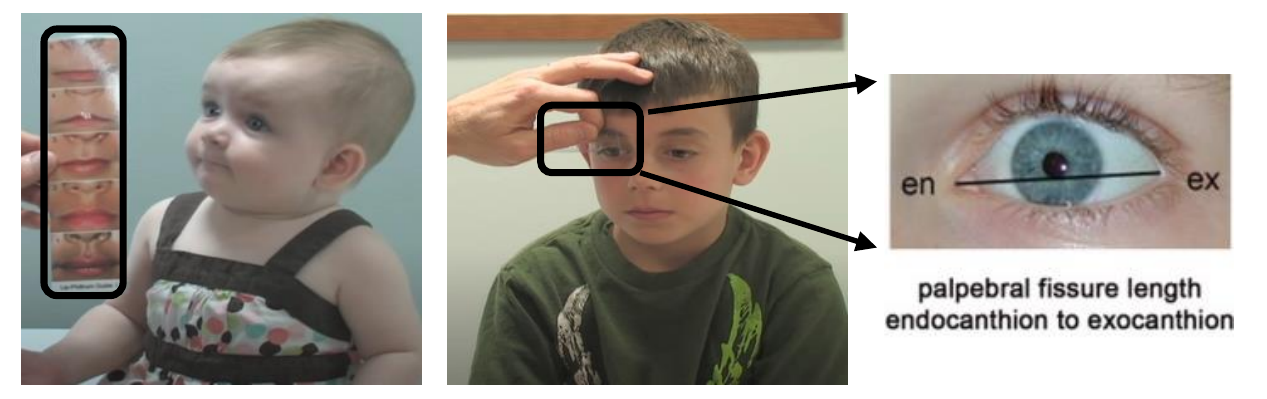}
\caption{Conventional clinical approaches for extracting diagnostic facial features of fetal alcohol syndrome (FAS). (Image credit: \cite{PFL}, \cite{youtube})}
\label{fas_illustration}
\end{figure}

\begin{figure}[t]
	\centering
	\setlength{\belowcaptionskip}{-0.5cm}
	\includegraphics[width=0.9\linewidth]{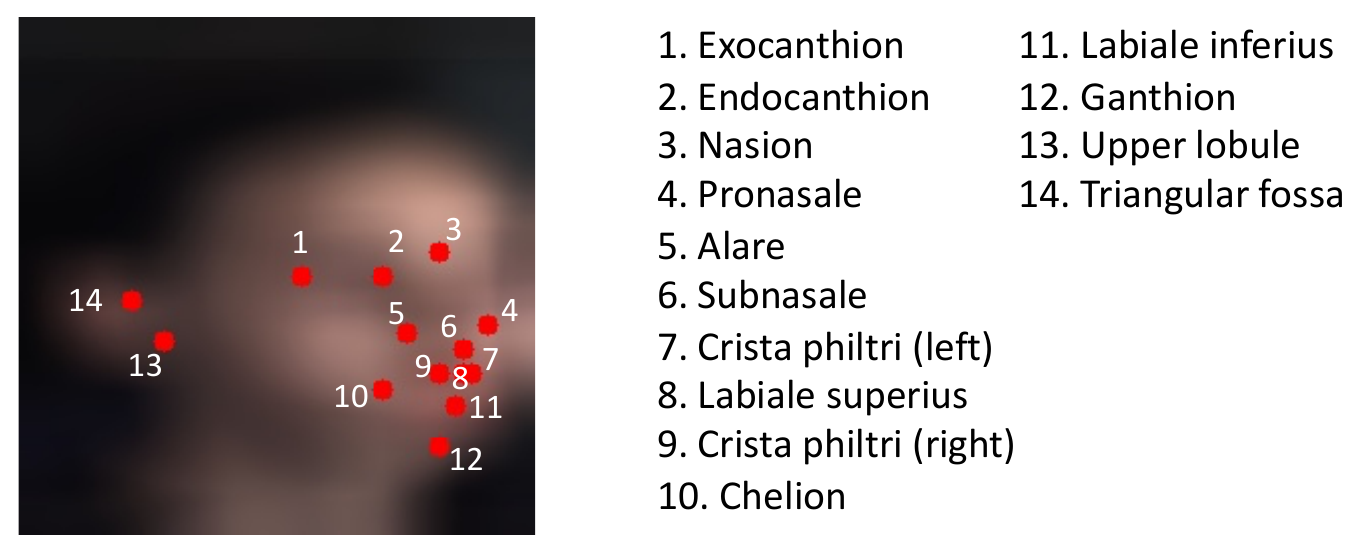}
	\caption{Illustration of anatomical landmarks used in this study.}
	\label{landmark_illustration}
\end{figure}

Conventional clinical approaches \cite{PFL,three-facial} for facial assessment rely on manual examination, using a ruler for PFL measurement, and a Likert-scale comparison chart for the lip and philtrum, as depicted in Fig. \ref{fas_illustration}. Clinical examination is necessarily subjective, with accuracy heavily reliant on the skill and experience of the clinician. Consequently, there is poor inter-rater reliability, and underdiagnosis and misdiagnosis of FASDs are a common occurrence \cite{three-facial}. Therefore, automatically localizing anatomical landmarks, as shown in Fig. \ref{landmark_illustration}, is seen as a crucial part of the clinical workflow to provide objective measurements for recognition of FASD.

\begin{figure}[t]
	\centering
	\setlength{\belowcaptionskip}{-0.5cm}
	\includegraphics[width=0.85\linewidth]{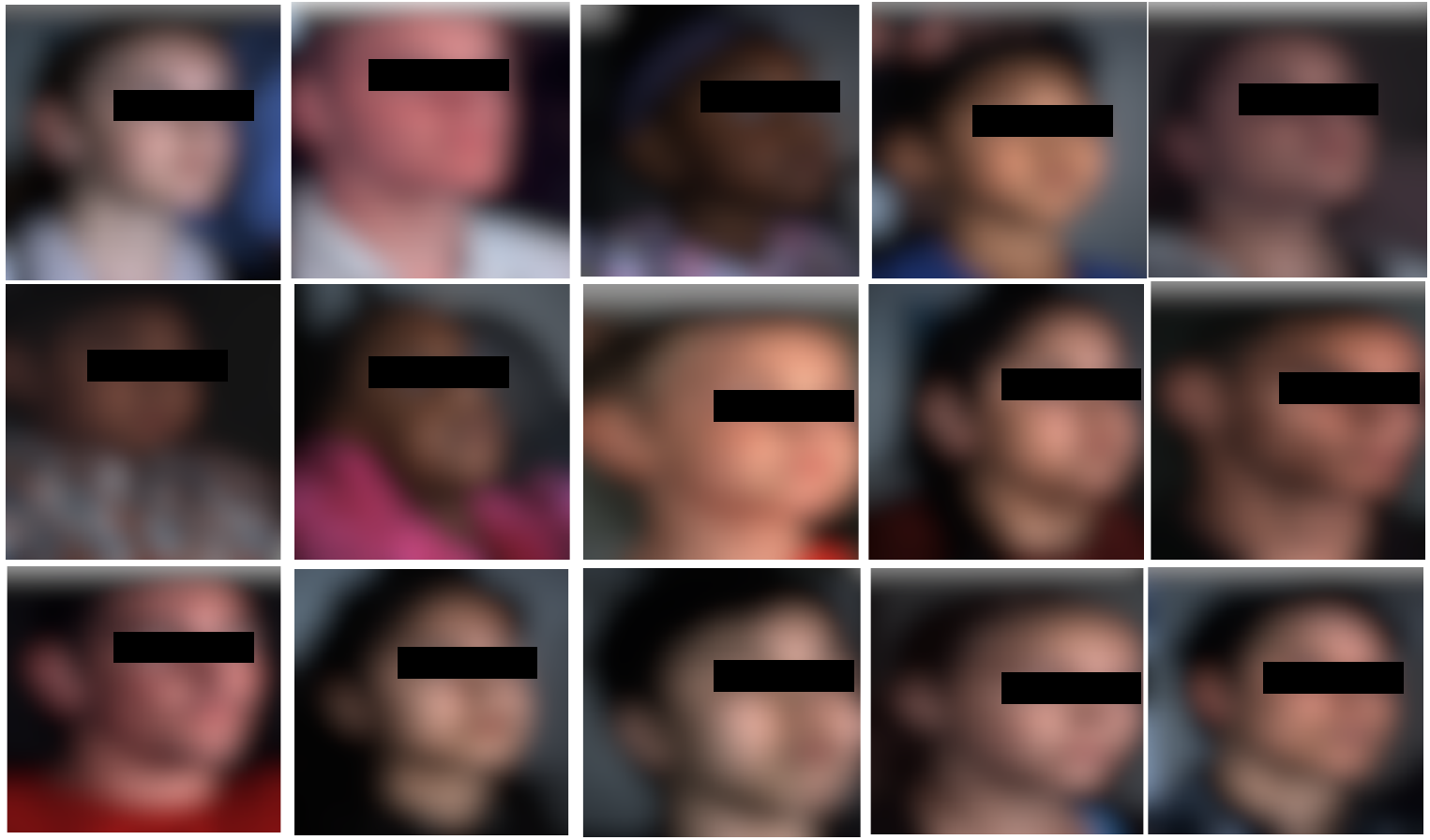}
	\caption{Examples of facial phenotypes in our challenging dataset.}
	\label{dataset_illustration}
\end{figure}

With the rapid development of deep learning, recent years have witnessed important advances in automation of various medical imaging tasks \cite{Unet}. Included in this, recent approaches  for anatomical landmark detection \cite{Deep-Anatomical-Cephalometric,regression-voting,Two-StageTask-Oriented, deep-landmark-medical, DSNT}, have often leveraged the power of deep neural networks, 
due to the ability to capture hierarchical representations from raw data, and have achieved promising performance.
However, anatomical landmark detection in our imaging application remains a challenge due to the large appearance variation of faces, different age groups and ethnic background, as exemplified in Fig. \ref{dataset_illustration}.
One feasible solution to mitigate the challenge is to utilize backbone CNN architectures (e.g. VGG16 \cite{vgg16} or ResNet50 \cite{resnet}) trained on a large-scale and diverse image dataset, such as VGGFace2 \cite{VGGFACE2}, which can be fine-tuned or specialized with additional task-specific layers to promote the optimization of facial anatomical landmark detection \cite{self-supervised,mobileFAN}.
Finetuning, as a common paradigm in transfer learning \cite{transfer-learning}, aims to benefit the target task by providing a good initialization, but it can require exhaustive tuning or a set of ad-hoc hyper-parameters to achieve good performance \cite{Inductive-TL,DELTA}. Furthermore, limited availability of labeled data, which often occurs in the medical domain, can result in overfitting in transfer learning, leading to poor accuracy for the desired task\cite{Two-StageTask-Oriented}.

\begin{figure}[t]
\centering
\setlength{\belowcaptionskip}{-0.3cm}
\includegraphics[width=0.95\linewidth]{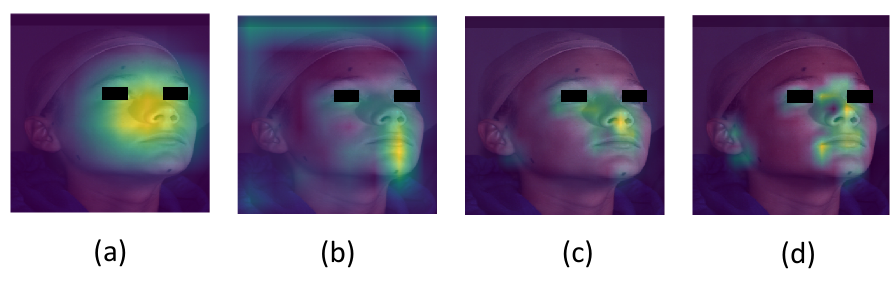}
\caption{Illustration of top-level activation attention maps for different approaches given an image in the target dataset, where (a) represents the original source model, (b) is the fine-tuned model with weight decay, (c) denotes the proposed regularization constraint on intermediate layer outputs, and (d) demonstrates the proposed regularization constraint on spatial attention maps.}
\label{justification_results}
\end{figure}

To address this, $L^{2}$ regularization, also known as weight decay, has been frequently suggested to reduce the effect of overfitting by regularizing the parameter search space \cite{Inductive-TL}. However, such a regularizer may not perform well in transfer learning, especially, and as in our case, when transferring knowledge across different tasks (classification $\rightarrow$ regression). We illustrate this in Fig. \ref{justification_results} (b) which shows that standard fine-tuning with weight decay ignores useful facial semantics or patterns (see Fig. \ref{justification_results} (a)) which are related to the target task of landmark detection, but instead can pay attention to irrelevant regions.
To this end, we argue that the target (fine-tuned) model should not only focus on adjusting the pre-trained weights, but also learn how to ``behave" as a source model does  \cite{DELTA},  so that facial semantics can be explicitly reused to improve learning of the landmark detector, as shown in Fig. \ref{justification_results} (c). 
Therefore, we propose to retain feature responses of a source classification model given the target task data to regularize  landmark detection optimization, as shown in Section \ref{Regularized Transfer Learning for Anatomical Landmark Detection}.
Inspired by knowledge distillation for model compression \cite{knowledge-distillation,hint-learning,attention-transfer}, we develop four regularization strategies during fine tuning, including constraining the feature outputs from classification and intermediate layers, as well as matching activation attention maps in both spatial and channel levels  which are shown to provide more comprehensive representations, as depicted in Fig. \ref{justification_results} (d).
Extensive experimental evaluations reported in Section \ref{Experiments}, demonstrate that the proposed framework generalizes well in anatomical landmark detection with limited training samples and favorably outperforms other solutions from the literature. 

This paper considerably extends our preliminary work in \cite{CTRL}, developing a unified regularized transfer learning framework for localizing anatomical landmarks in fetal alcohol syndrome. The new contributions of this paper relative to \cite{CTRL} are, 
1)  we present a more detailed justification of the proposed framework  as well as its importance with additional illustrations.
2) we propose two additional regularization constraints based on spatial and channel attention maps which improve localization accuracy relative to results reported in \cite{CTRL}. The technical details of each solution in the proposed framework are explained.
3) Experimentally, we provide an in-depth analysis of different proposed regularization strategies, analyze the effectiveness of different pre-trained source models for the proposed framework,  study the impact of different parameter selection and include detailed comparisons with further related methods both quantitatively and qualitatively. 
4) we provide a detailed discussion to discuss the strengths and the limitations of our approaches and suggest their translational potentials.

The rest of the paper is structured as following: Section \ref{Related-Work} discusses the most closely related work to the proposed approach. Section \ref{Method} formulates the problem considered and details the proposed solution via regularized transfer learning. In Section \ref{Experiments}, we perform extensive experimental evaluations. Finally, following the Discussion in Section \ref{Discussion}, Section \ref{Conclusions} draws conclusions and suggests some future directions of research.
\section{Related Work}
\label{Related-Work}
\subsection{Anatomical Landmark Detection}
In the literature, there has been significant progress on automatic landmark (keypoint) detection of natural images, which has in turn advanced human pose estimation, and facial alignment.
Existing methods for this task include  random forest \cite{regression-forest,geometric-configuration} and cascaded shape regression \cite{explicit-regression,cascaded-pose}.
Convolutional neural networks (CNN) based regression has become a recently popular approach for keypoint localization \cite{stacked-hourglass,mobileFAN,wing-loss,high-resolution}.
Two CNN-based architecture designs have emerged: direct coordinate regression \cite{facial-attribute,wing-loss} and heatmap regression \cite{stacked-hourglass,simple-baseline}, where the latter usually outperforms the former, due to the advantage of preserving higher spatial resolution for accurate localization. Methods based on U-Net \cite{Unet}, a stacked hourglass \cite{stacked-hourglass}, and simple-baseline \cite{simple-baseline} CNN architectures have achieved good performance on the challenging keypoint detection benchmarks (e.g. MPII benchmark \cite{mpii}).

Different from generic keypoint localization in natural images,
anatomical landmark detection in medical imaging is often constrained by the scarcity of labeled data, as annotating abundant landmarks in the medical domain is expensive.
To accommodate this, some existing methods have considered cascaded CNNs following a coarse-to-fine strategy \cite{deep-landmark-medical,Two-Stage-Task-Oriented,spatial-configuration}.
Although cascaded designs can give good results,  they require additional computational efforts, where repeatedly extracting similar low-level features may not be necessary \cite{Attention-UNet}.
On the other hand, attention mechanisms which were originally proposed for machine translations \cite{attention-need}, have been exploited to enhance the CNN-based anatomical landmark detection \cite{attention-guided-landmark,regression-voting,Misshapen-Pelvis-Landmark}.
Specifically, Zhong et al. \cite{attention-guided-landmark} developed a two-stage U-Net with attention guidance to localize anatomical landmarks. Chen et al. \cite{regression-voting} and Liu et al. \cite{Misshapen-Pelvis-Landmark} exploited self-attention modules \cite{attention-need},\cite{non-local} for pelvis and cephalometric landmark detection respectively.
More recently, Huang et al. \cite{DSNT} added a differentiable spatial to numerical transform layer to an encoder-decoder network.
Vlontzos et al. \cite{regression-DQN} introduced multi-agent reinforcement learning for multiple landmark detection  by designing a collaborative Deep Q-Network.
Unlike the aforementioned methods, the proposed learning approach focuses on internally enriching the feature representations via regularized transfer learning for anatomical landmark detection, requiring neither attention mechanisms nor cascaded processing.

\subsection{Knowledge Transfer}
Fine-tuning \cite{simple-baseline} is a standard practice in transfer learning for refining a pre-trained model to be suitable for a new target task. Typically a small learning rate is applied and some model parameters may need to be frozen to reduce overfitting.
However, empirically modifying existing model parameters is not a direct method to reuse the representational features from a source task model, which may limit good generalization to a small new task training dataset \cite{LWF}.

In contrast, knowledge distillation, originally proposed for model compression \cite{knowledge-distillation},  can be considered as a more direct way to retain the learning experiences from a source task model by  encouraging predictions of a student model to be close to those of a teacher model.
This technique has been extended and applied to various applications, including hint learning \cite{hint-learning}, incremental learning \cite{LWF,LWM,Incremental-ultrasound}, privileged learning \cite{privileged-info}, domain adaptation  \cite{cross-modality} and human expert knowledge distillation \cite{gaze-distillation}.
Distillation methods focus on producing a compact model by operating the knowledge transfer across common tasks \cite{knowledge-distillation,gaze-distillation,hint-learning}.
Different from the aforementioned methods, the proposed learning approach in this paper regularizes the transfer learning across different tasks.
In other words, the proposed regularized knowledge transfer can be interpreted as cross-task representation learning, where ``cross-task" means that the learning process is made between the source and target tasks corresponding to a different label space.

To boost the performance of the target task, joint training of the source and the target tasks, termed multi-task learning \cite{facial-attribute} has been proposed, where model parameters for both tasks are jointly optimized to learn a shared representation to benefit the target task. However that method requires data and labels of both tasks to be available during training, which can be difficult to satisfy for modelling tasks from clinical data.
For our proposed design, we do not require access to the data from the source domain.

\begin{figure*}[t]
\centering
\setlength{\abovecaptionskip}{-3pt}
\setlength{\belowcaptionskip}{-5pt}
\includegraphics[width=0.85\linewidth]{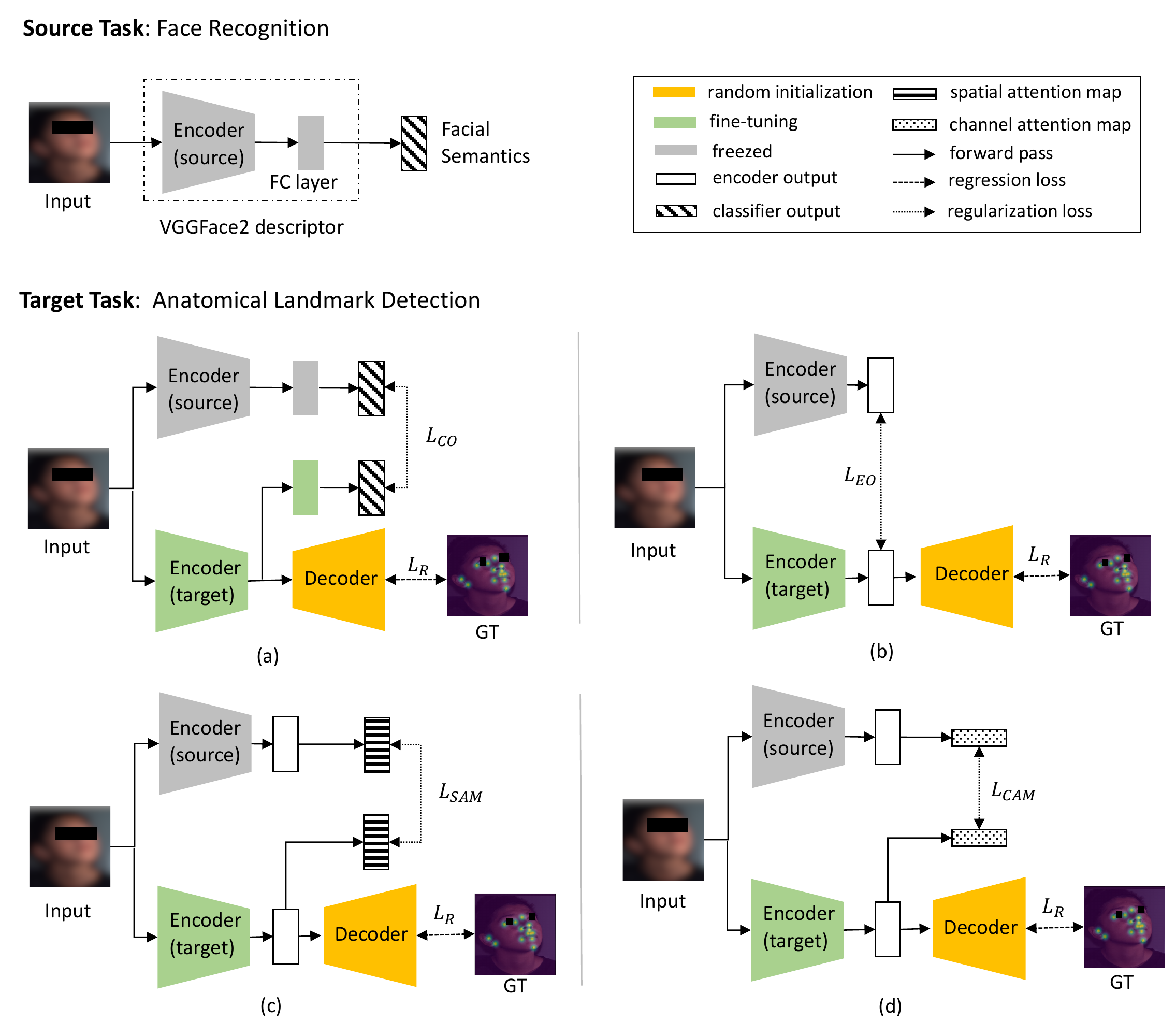}
\caption{Illustration of proposed regularized transfer learning for the anatomical landmark detection. Face recognition is considered as the source task to provide rich facial semantics. For the target task, (a)-(d) illustrate four regularization constraints on the classification output ($L_{CO}$), the encoder output ($L_{EO}$), the spatial attention map ($L_{SAM}$) and the channel attention map ($L_{CAM}$), respectively.}
\label{overall-system}
\end{figure*}
\section{Method}
\label{Method}
In this section, we first introduce the task of anatomical landmark detection, then describe the proposed regularized transfer learning solution for this task.
\subsection{Problem Formulation}
\label{problem-formulation}
Consider a supervised anatomical landmark detection task,
where we are given a training dataset with $N$ pairs of training samples, denoted as $\mathcal{D}= \{\mathbf{I}_{i}, \mathbf{p}_{i}\}_{i=1}^{N}$, where $\mathbf{I}_{i}\in \mathbb{R}^{H \times W \times 3}$ represents a 2D facial image with height $H$ and width $W$, and 
$\mathbf{p}_{i} = [(x_{1}, y_{1}), (x_{2}, y_{2}),...,(x_{K}, y_{K})] \in \mathbb{R}^{2\times K}$ denotes $K$ numbered labeled landmark coordinates in each of the N images ($i=1 \rightarrow N$).

Inspired by its recent success in keypoint localization \cite{stacked-hourglass,simple-baseline}, we use a heatmap to regress the keypoints.
To do this, we need to transform these landmark coordinates into heatmaps.  Specifically, labeled coordinates are downscaled to $1/4$ of the input size ($\mathbf{p}_{i} \leftarrow \mathbf{p}_{i}/4$), and then are transformed to a set of heatmaps $\mathbf{G}_{i} \in \mathbb{R}^{(H/4) \times (W/4) \times K}$. Each heatmap $\mathbf{g}_{i, k} \in \mathbb{R}^{(H/4) \times (W/4)}, k\in \{1,...,K\}$ is defined as a 2D Gaussian kernel centered on the $k$-th landmark coordinate $(x_{i, k},y_{i, k})$, which is given as,
\begin{equation}\label{heatmap}
\mathbf{g}_{i, k}(a,b)=\exp \left(-\frac{(a-x_{i, k})^{2}+(b-y_{i, k})^{2}}{2 \sigma^{2}}\right)
\end{equation}
where $a$ and $b$ denotes the entry of matrix $\mathbf{g}_{i, k}$ in the $a$-th row and $b$-th column, and $\sigma$ is Gaussian kernel scale parameter.

Given the dataset $\mathcal{D}= \{\mathbf{I}_{i}, \mathbf{G}_{i}\}_{i=1}^{N}$, the goal is then to learn a non-linear mapping function between an input image and a set of heatmaps.
We follow  \cite{simple-baseline,mobileFAN}, and design a CNN-based encoder-decoder network $\mathcal{T}_{\theta_{e},\theta_{d}}: \mathbb{R}^{H \times W \times 3}\rightarrow \mathbb{R}^{(H/4) \times (W/4) \times K}$ with learnable parameters $\theta_{e}$ and $\theta_{d}$, to achieve the goal.
To mitigate against facial variability in our dataset, we consider a ResNet-50 model \cite{resnet} pre-trained on a large-scale and
diverse VGGFace2 \cite{VGGFACE2} dataset as a backbone for feature extraction (encoder) $f_{\theta_{e}}: \mathbb{R}^{H \times W \times 3}\rightarrow \mathbb{R}^{m}$, with the classification branch $g_{\theta_{c}}: \mathbb{R}^{m} \rightarrow \mathbb{R}^{C} $ being disabled, where $m$ denotes the dimensionality of the encoder output, and $C$ represents the number of classes in the source classification task.
For the decoder, three deconvolutional layers are employed to recover the spatial resolution, each of which is with the dimension of 256, a $4 \times4$ kernel and a stride of $2$, and is followed by a batch normalization and a rectified linear unit (ReLU) \cite{relu} activation function. Then, a $1 \times 1$ convolutional layer with linear activation is applied to obtain the predicted heatmaps $h_{\theta_{d}}: \mathbb{R}^{m}\rightarrow \mathbb{R}^{(H/4) \times (W/4) \times K}$.
Consequently, the network $\mathcal{T}_{\theta_{e},\theta_{d}} := h_{\theta_{d}}(f_{\theta_{e}}(\mathbf{I}))$ can be learned by minimizing the following loss between the predicted heatmaps and the corresponding ground truth,
\begin{equation}\label{regression-loss}
L_{R}= \frac{1}{N}\sum_{i=1}^{N} \left\|\mathbf{G}_{i}-h_{\theta_{d}}(f_{\theta_{e}}(\mathbf{I}_{i}))\right\|^{2}_{F}
\end{equation}
where $F$ denotes the Frobenius norm.

The above underpins our primary learning objective for the task of anatomical landmark detection. Standard training strategies for such a  transfer learning approach usually rely on fine-tuning, where the entire network or only the decoder is trained on the target data.
Considering the limited available labelled data in our case, an L2 penalty (weight decay) is usually utilized in the fine-tuning to reduce parameter over-fitting. However, merely penalizing the model weights may not  capture the original knowledge embedded in the source network. In our case this may lead to a loss of useful facial descriptions and the distraction from other irrelevant regions, as illustrated in Fig. \ref{justification_results}.
We present proposed solutions to address this problem in the next subsection.
\subsection{Regularized Transfer Learning for Anatomical Landmark Detection}
\label{Regularized Transfer Learning for Anatomical Landmark Detection}
In order to reuse knowledge from the source network in the landmark detector model, we introduce a new regularized transfer learning (RTL) approach which aligns the network features (behaviors) of the source and the target models during target model training.
To this end, we compute predictions of the source network on the target dataset $\mathcal{D}^{t}$, which are matched with those from the target network through a regularization loss $L_{\gamma}$.
Hence, the total loss is updated as,
\begin{equation}\label{total-loss}
L=L_{R}+ \lambda L_{\gamma}
\end{equation}
where $\lambda>0 $ is a weighting parameter. If $\lambda = 0$, Eq. (\ref{total-loss}) reduces to standard fine-tuning. Fig. \ref{overall-system}  shows the overall architecture of proposed regularized transfer learning for anatomical landmark detection. Face recognition is considered as the source task to provide rich facial semantics. For the target task, as shown in Fig. \ref{overall-system} (a)-(d), there are four different regularization constraints that are developed and explored for the proposed learning approach. 
\subsubsection{Constraint on Classifier Output (RTL-CO)}
Motivated by  \cite{knowledge-distillation} \cite{LWF}, this variant constrains the distance between the classification outputs for the design of $L_{\gamma}$, as shown in Fig. \ref{overall-system} (a). To do this, the previously abandoned classification branch $g_{\theta_{c}}: \mathbb{R}^{m} \rightarrow \mathbb{R}^{C} $ in Section \ref{problem-formulation} is re-enabled for training. Similar to the distillation loss \cite{knowledge-distillation}, we use a temperature parameter $\mu$ with $softmax$ function denoted as $\sigma$ to smooth the predictions. While the original cross-entropy function is replaced by the following term,
\begin{equation}\label{output-loss}
L_{CO}= \frac{1}{N}\sum_{i=1}^{N}\left\|\sigma\left( \frac{g_{\theta_{c}}^{s}(f_{\theta_{e}}^{s}(\mathbf{I}_{i}))}{\mu}\right)-\sigma\left( \frac{g_{\theta_{c}}^{t}(f_{\theta_{e}}^{t}(\mathbf{I}_{i}))}{\mu}\right)\right\|^{2}_{2}
\end{equation}
The main reason for this replacement is that the task (label space) between the source and target domains is different, and applying cross-entropy function as a strong regularizer could make the optimization lean towards the source task, thereby affecting the target task performance. Moreover, our empirical results showed that using the L2 norm improves the cross-entropy function by 0.45 pixels regarding the mean square errors.
Note that the superscripts $s$ and $t$ refer to the source and target (fine-tuned) networks respectively.
\subsubsection{Constraint on Encoder Output (RTL-EO)}
this variant considers the role of  the regularization constraint, as shown in Fig. \ref{overall-system} (b). The goal is to align the features from encoder outputs, similar to hint learning \cite{hint-learning}.
Compared to  $L_{CO}$, aligning the intermediate feature maps which preserve spatial configuration can be more effective and helpful for the landmark localization \cite{cross-modality}. 
In this case, we adopt the cosine similarity for feature alignment as described in Eq. (\ref{encoder-loss}) \cite{RKD}, as we conjecture that penalizing higher-order angular differences would benefit matching the relational information between the source and target tasks. 
\begin{equation}\label{encoder-loss}
L_{EO}= 1-  \frac{f_{\theta_{e}}^{s}(\mathbf{I}_{i})\cdot f_{\theta_{e}}^{t}(\mathbf{I}_{i})}{\| f_{\theta_{e}}^{s}(\mathbf{I}_{i})\| \times \|f_{\theta_{e}}^{t}(\mathbf{I}_{i})\|}
\end{equation}

\subsubsection{Constraint on Spatial Attention Map (RTL-SAM)}
In addition to purely matching the predicted scores or feature maps between source and target tasks, it is important to understand how retainment of source task facial knowledge is beneficial to the target task. In this regard, we propose to constrain spatial attention maps which indicate where the meaningful features are emphasized given the input data. 
This is inspired by  \cite{attention-transfer}, \cite{block-attention} which demonstrated attention maps can be more interpretative than estimated feature maps.  Fig. \ref{justification_results} (c) and Fig. \ref{justification_results} (d) demonstrate the advantage of \textit{RTL-SAM} over that of \textit{RTL-EO }on an example image.
We specifically employ activation-based spatial attention maps \cite{attention-transfer} for the  proposed \textit{RTL-SAM}, as shown in Fig. \ref{overall-system} (c).
Given an activation map $\mathbf{E}_{i} \in \mathbb{R}^{H \times W \times B} = f_{\theta_{e}}(\mathbf{I}_{i})$ for the $i$-th sample of the target dataset, the corresponding spatial attention map $\mathbf{A}_{i} \in \mathbb{R}^{H \times W}$ is computed as \cite{attention-transfer},
\begin{equation}\label{activation-attention}
  \mathbf{A}_{i} =\sum_{b=1}^{B}\left|\mathbf{E}_{i}(b)\right|^{2}
\end{equation}
where $B$ denotes the number of feature channels, $|\cdot|$ and $(\cdot)^{2}$ are the element-wise absolute value and power operators, respectively.
We then reshape the $\mathbf{A}_{i}\in \mathbb{R}^{H \times W}$ into $\hat{\mathbf{A}}_{i}\in \mathbb{R}^{H W}$ and use it to formulate the following loss function \cite{attention-transfer},
\begin{equation}\label{SAM-loss}
L_{SAM}= \frac{1}{N}\sum_{i=1}^{N}\left\|
\frac{\hat{\mathbf{A}}_{i}^{s}}{\|\hat{\mathbf{A}}_{i}^{s}\|_{2}}-
\frac{\hat{\mathbf{A}}_{i}^{t}}{\|\hat{\mathbf{A}}_{i}^{t}\|_{2}}\right\|^{2}_{2}
\end{equation}
\subsubsection{Constraint on Channel Attention Map (RTL-CAM)}
The final variant considers channel-wise attention maps \cite{block-attention,SE-Net} as  a regularization constraint in our proposed framework. See Fig. \ref{overall-system} (d).
Here we squeeze the spatial dimension of an activation map $\mathbf{E}_{i} \in \mathbb{R}^{H \times W \times B} $ to achieve a context vector $\mathbf{Q}_{i} \in \mathbb{R}^{B}$ \cite{SE-Net}, as given below,
\begin{equation}
  \mathbf{Q}_{i} = \frac{1}{H \times W}\sum^{H}_{h=1}\sum^{W}_{w=1}|\mathbf{E}_{i}(h,w)|
\end{equation}
Similar to $L_{SAM}$, we formulate the regularization loss $L_{CAM}$ as below,
\begin{equation}\label{CAM-loss}
L_{CAM}= \frac{1}{N}\sum_{i=1}^{N}\left\|
\frac{\mathbf{Q}_{i}^{s}}{\|\mathbf{Q}_{i}^{s}\|_{2}}-
\frac{\mathbf{Q}_{i}^{t}}{\|\mathbf{Q}_{i}^{t}\|_{2}}\right\|^{2}_{2}
\end{equation}
%

In practice, the four regularization terms $L_{CO}$, $L_{EO}$, $L_{SAM}$  and $L_{CAM}$ can be combined in the total loss function as described in \ref{total-loss}. The effect on performance of doing this will be assessed in the following section.
For the model inference, only the trained target encoder and decoder are utilized to predict heatmaps $\mathbf{H}_{i} = h_{\theta_{d}}^{t}(f_{\theta_{e}}^{t}(\mathbf{I}_{i}))$, and they are further processed via an $argmax$ function to obtain final landmark locations $\hat{\mathbf{p}}_{i}$. Note that other methods such as fitting another radial basis function to obtain subpixel localization can be also utilized in here for kepoint extraction.
%
\section{Experiments}
\label{Experiments}
\subsection{Dataset}
\begin{table*}[t]
	\caption{Quantitative comparison between proposed and other related existing approaches using MSE$\pm$Std and failure rate on the test set. Best results are marked in bold.}
	\centering
	\label{table1}
	\resizebox{0.75\linewidth}{!}{%
		\begin{tabular}{c|c|c|c|c|c|c|c}
			\toprule
			\multirow{2}{*}{Method} & \multirow{2}{*}{\begin{tabular}[c]{@{}c@{}}MSE$\pm$Std\\ ($\downarrow$)\end{tabular}} & \multicolumn{3}{c|}{Failure Rate ($\downarrow$)} & \multicolumn{3}{c}{AUC ($\uparrow$)} \\ \cline{3-8}
			&  & 1.0 pixel & 1.2 pixel & 1.4 pixel&1.0 pixel&1.2 pixel&1.4 pixel \\ \hline
			FE \cite{simple-baseline}& 1.82$\pm$0.50 & 99.03\% & 94.52\% &80.00\%&0.01&0.01&0.02 \\
			DSNT \cite{DSNT}&1.41$\pm$0.26 & 96.13\%& 74.19\%& 50.65\%&0.01&0.02&0.07\\
			FTP \cite{simple-baseline} & 1.16$\pm$0.26 & 73.87\% & 40.32\% &16.77\% &0.04&0.10& 0.19\\
			U-Net \cite{Unet}&0.93$\pm$0.39 &30.00\% & 10.00\%& 4.84\%&0.15&0.26&0.36 \\
			HG \cite{stacked-hourglass}& 0.88$\pm$ 0.39&24.84\% & 12.58\%&6.77\%&0.20&0.30&0.39 \\
			FT \cite{simple-baseline}& 0.86$\pm$0.24 &24.19\%  & 10.65\% &3.23\%  &0.18&0.29&0.38\\ \hline
			\emph{RTL-CO} & 0.84$\pm$0.25 & 24.84\% &  5.81\% & 1.61\% &0.20&0.31&0.40\\
			\emph{RTL-EO} & 0.83$\pm$0.25 & 21.29\% &  7.74\%& 1.61\% &0.21&0.32&0.41\\
			
			\emph{RTL-SAM} & \textbf{0.78$\pm$0.22} & \textbf{13.55\% }&  \textbf{4.84\%}& \textbf{1.61\%} &\textbf{0.24}&\textbf{0.35}&\textbf{0.44}\\
			\emph{RTL-CAM} & 0.82$\pm$0.24 & 20.00\% &  5.81\%& 1.61\% &0.21&0.32&0.42\\
			\bottomrule
		\end{tabular}%
	}
	
\end{table*}

\label{Dataset}

The data used in the experiments was collected by the Collaborative Initiative on Fetal Alcohol Spectrum Disorders (CIFASD)\footnote{https://cifasd.org/}, a global multi-disciplinary consortium focused on furthering the understanding of FASD. 
It contains subjects from 4 sites across the USA, aged between 4 and 18 years. Each subject was imaged using a commercially available static 3D photogrammetry system from 3dMD\footnote{http://www.3dmd.com/}. 
Since directly processing a high spatial resolution 3D mesh is computationally intensive, we utilize the 2D images captured during 3D acquisition, which are used as UV mapped textures for 3D surface reconstruction.
We used in total 1549 facial scans for our experiments, and for each scan, there are 14 FAS associated landmarks which had been manually annotated by an expert (see Fig. \ref{landmark_illustration} for details). These  scans were randomly split into a training/validation set (70\%/10\%), and a test set (20\%) for the experiments.

\subsection{Implementation Details}
\begin{figure}[H]
	\centering
	\includegraphics[width=1\linewidth]{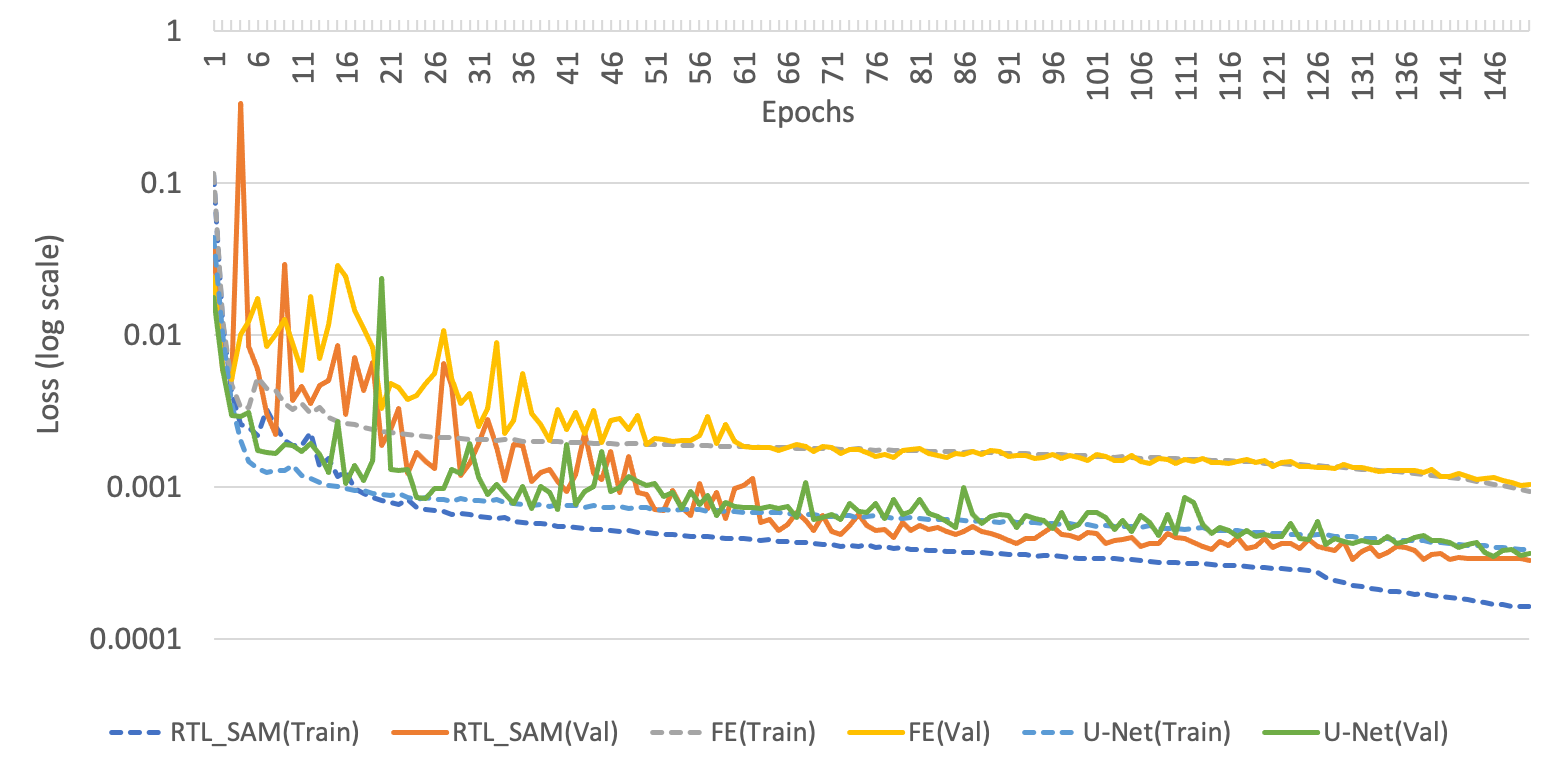}
	\caption{Illustration of training and validation loss curve.}
	\label{tran_val_loss}
\end{figure}

All the images were cropped and resized to $256 \times 256$ pixels, and intensity normalized with mean and standard deviation (std) values provided by the VGGFace2 \cite{VGGFACE2} for the network training, validation and testing.
Data augmentation was performed with randomly horizontal flip (50\%) and scaling ($0.8$).
During training, the Adam optimizer \cite{Adam} was used for the optimization ($\beta_{1} = 0.9$, $\beta_{2}=0.999$, and epsilon = $1e^{-8}$) with a mini-batch size of $2$ for $150$ epochs. A polynomial decay learning rate was used with an initial value of $0.001$.  Models saved with the best accuracy on the validation set were used for evaluation. For more details of model training, Fig. \ref{tran_val_loss} in Section \ref{Visualization of Training Details} depicts a learning curve of model performance on the train/validation sets.
The temperature parameter $\mu$ in Eq. (\ref{output-loss}) was set to $2$. We experimentally set up the weighting parameter $\lambda=0.002$ used in Eq. (\ref{total-loss}) and the kernel scale parameter $\sigma = 1.5$ used in Eq. (\ref{heatmap}), and we used this parameter setting throughout the remaining experiments.
The proposed learning algorithm was implemented using TensorFlow 1.6.0 and the model trained using an Nvidia GTX 1060 GPU.

\subsection{Evaluation Metrics}
\begin{figure*}[t]
	\centering
	\includegraphics[width=0.95\linewidth]{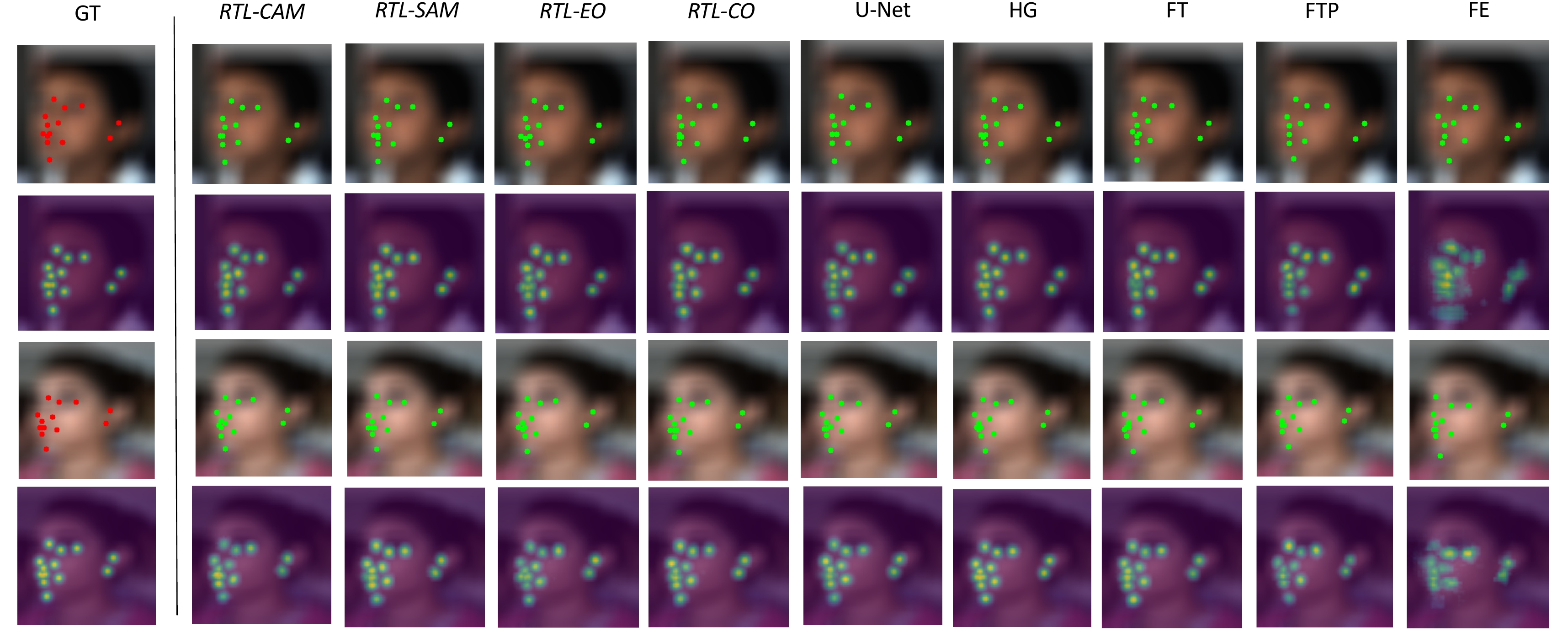}
	\caption{Qualitative evaluation of landmark detection (first and third rows) and heatmap regression (second and fourth rows) on the test set.  Best viewed in color version.}
	\label{qualitative_results}
\end{figure*}

We firstly employ the Mean Squared Error (MSE) for evaluation, which is a commonly-used metric to assess landmark detection. It is defined as:
\begin{equation}\label{evaluation-metric}
\mathrm{MSE}=\frac{1}{N_{e}} \sum_{i=1}^{N_{e}} \frac{1}{K}\left\|\mathbf{p}_{i}-\hat{\mathbf{p}}_{i}\right\|_{2}
\end{equation}
where $N_{e}$ is the number of images in the test set, and $\mathbf{p}_{i}$ and $\hat{\mathbf{p}}_{i}$ denote the ground truth and prediction landmark, respectively.
Note that the original normalization factor estimated by inter-ocular distance (Euclidean distance between outer eye corners) \cite{300W} is not included in this evaluation, as annotations for the other eye are not available, as shown in Fig. \ref{qualitative_results}.
In addition, we utilize the Cumulative Errors Distribution (CED)
curve with two additional metrics of Area-Under-the-Curve (AUC) and Failure Rate (FR).
When the Euclidean distance between the estimated landmark and its ground truth location (point-to-point Euclidean error) is greater than a pre-defined threshold, it is considered as a mis-detection.  We consider three thresholds of 1.0 pixels, 1.2 pixels and 1.4 pixels.
%
\begin{figure}[t]
	\centering
	\includegraphics[clip, trim=0.4cm 0.01cm 2.cm 3.cm,width=0.8\linewidth]{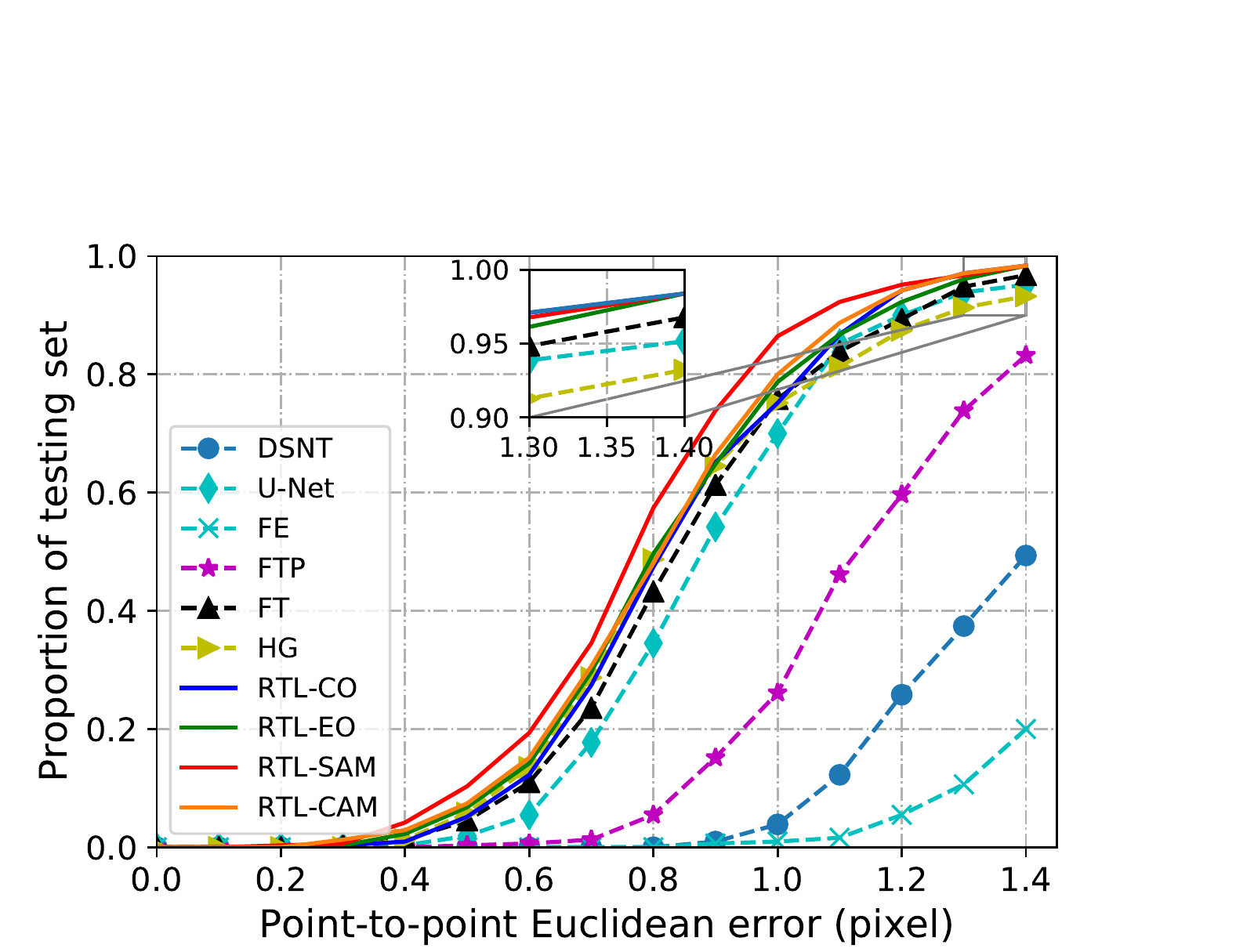}
	\caption{Illustration of cumulative error distribution (CED) curve on the test set. Best viewed in colored version.}
	\label{ced-results}
\end{figure}
\subsection{Comparison with Existing Approaches}
\label{Comparison with Existing Approaches}
In this section, we evaluate the effect of the proposed regularized transfer learning (RTL) methods (\emph{RTL-CO}, \emph{RTL-EO}, \emph{RTL-SAM}, and \emph{RTL-CAM}) and compare these variants with several related approaches from the literature. The related approaches are: a two-stage Hourglass network (HG) \cite{stacked-hourglass}, a modified U-Net for landmark localization \cite{Unet}, and a fully-convolutional network with a differentiable spatial-to-numerical transform (DSNT) \cite{DSNT}. In addition, three variants of transfer learning  \cite{simple-baseline} without regularization ($\lambda = 0$) are considered: Feature Extraction (FE) \cite{simple-baseline}  with freezing the encoder, Fine Tuning Parts (FTP) \cite{simple-baseline}  without freezing the final convolutional layer of the encoder, and Fine Tuning (FT) \cite{simple-baseline}  without freezing any layer. For a fair comparison, all the evaluated methods use the same weight decay rate of $10^{-4}$.

Table \ref{table1} summarizes the quantitative evaluation on the test set by reporting the MSE, failure rate, and AUC for each compared method. Performance metrics with ($\downarrow$) or ($\uparrow$) indicate that lower is better or higher is better, respectively.
From the results, we observe that all the proposed new approaches favorably surpass the previous methods from the literature for most metrics.
It is noteworthy that  \emph{RTL-SAM} reports the best performance among all the compared approaches.
Compared to FT \cite{simple-baseline}, which is closely related to our work, the proposed \emph{RTL-SAM} improves the MSE by 0.08 pixels (9.3\%), and the failure rate by 11.29\%, 5.81\%, and 1.62\% for the three thresholds. This demonstrates that the proposed regularization constraint is beneficial for transfer learning across domain-similar source/target tasks.
\begin{table*}[t]
	\centering
	\caption{Ablation studies on different proposed regularization constraints. The upper part shows the results for the individual landmark localization using  MSE$\pm$  Std (pixel), while the lower part presents the failure rate with the error threshold of 1.2 pixel. Best results are shown in bold.}
	\label{table2}
	\resizebox{0.85\linewidth}{!}{
		\begin{tabular}{c|ccccccc}
			\toprule
			\multirow{2}{*}{Landmark} & \multicolumn{7}{c}{Configurations} \\ \cline{2-8}
			& \emph{RTL-CO} &\emph{ RTL-EO} & \emph{RTL-SAM} & \emph{RTL-CAM }& \emph{RTL-CO-EO} &\emph{ RTL-CO-SAM}& \emph{RTL-CO-CAM}\\ \hline
			L1 & 0.64$\pm$0.58  & 0.63$\pm$0.59  &\textbf{0.59$\pm$0.56}    &0.65$\pm$0.58 & 0.64$\pm$0.58  &0.61$\pm$0.60 & 0.62$\pm$0.56   \\
			L2 & 0.68$\pm$0.54  & 0.74$\pm$0.54 & \textbf{0.67$\pm$0.56}  &0.78$\pm$0.55  & 0.79$\pm$0.55 &0.74$\pm$0.58  & 0.77$\pm$0.64\\
			L3 & 0.82$\pm$0.63 & 0.89$\pm$0.60 & 0.74$\pm$0.59  &0.77$\pm$0.84  & 0.82$\pm$1.12 &0.75$\pm$0.60  & \textbf{0.73$\pm$0.72} \\
			L4 & 0.65$\pm$0.60 &\textbf{0.61$\pm$0.56} &0.66$\pm$0.93   &0.68$\pm$0.55 &0.62$\pm$0.56 &0.63$\pm$0.57  &0.64$\pm$0.58 \\
			L5 &0.71$\pm$0.57   & \textbf{0.64$\pm$0.56}&0.67$\pm$0.55   &0.69$\pm$0.56  &0.67$\pm$0.58  &0.66$\pm$0.57 & 0.73$\pm$0.56 \\
			L6 & 0.75$\pm$0.58  & 0.74$\pm$0.57 &\textbf{0.68}$\pm$\textbf{0.56}  &0.69$\pm$0.56  &0.73$\pm$0.56 &0.78$\pm$0.56  &0.79$\pm$0.58 \\
			L7 &1.48$\pm$1.17 & 1.46$\pm$1.13 &\textbf{1.20$\pm$1.02}   &1.46$\pm$1.22 &1.44$\pm$1.10  &1.58$\pm$1.29 &1.28$\pm$1.05  \\
			L8 &0.74$\pm$1.02 & 0.72$\pm$0.56 &0.73$\pm$0.56   &0.78$\pm$0.56  &0.74$\pm$0.55  &0.74$\pm$0.55 &\textbf{0.68$\pm$0.56} \\
			L9 &1.40$\pm$1.05 & 1.39$\pm$1.15 & \textbf{1.17$\pm$1.19}  &1.28$\pm$1.17  &1.30$\pm$1.19   &1.50$\pm$1.33  & 1.32$\pm$1.13\\
			L10 &0.64$\pm$0.57  & \textbf{0.60$\pm$0.55} &0.63$\pm$0.59   &0.60$\pm$0.58  &0.64$\pm$0.58  &0.63$\pm$0.57  &0.71$\pm$0.57 \\
			L11 &0.89$\pm$1.06  & \textbf{0.75$\pm$0.60} &0.80$\pm$0.58 &0.84$\pm$0.60  &0.83$\pm$0.61 &0.83$\pm$0.57 & 0.91$\pm$1.07 \\
			L12 &0.89$\pm$0.72  & 0.92$\pm$1.35 &0.85$\pm$0.62   &0.84$\pm$0.66  &\textbf{0.82$\pm$0.64}  &0.95$\pm$0.69  &0.94$\pm$0.63 \\
			L13 &0.72$\pm$0.60  & 0.70$\pm$0.61 &0.74$\pm$0.60   &\textbf{0.63$\pm$0.59}  &0.71$\pm$0.62  &0.69$\pm$0.60  &0.76$\pm$0.60 \\
			L14 &0.78$\pm$0.59  & 0.86$\pm$0.72 & 0.78$\pm$0.58  &0.80$\pm$0.63  &0.85$\pm$0.62  &\textbf{0.76$\pm$0.57}  &0.86$\pm$0.60 \\ \hline
			Average &0.84$\pm$0.25  &0.83$\pm$0.24  & \textbf{0.78$\pm$0.22}& 0.82$\pm$0.24 &0.83$\pm$0.25   & 0.85$\pm$0.25  &0.86$\pm$0.33  \\ \bottomrule
		\end{tabular}%
	}
\end{table*}

In Fig. \ref{ced-results}, we show  the cumulative distributions of landmark detection errors for the methods considered. This reveals the advantage of the proposed approaches over the compared previous ones, which is consistent with the findings in Table \ref{table1}.
From the CED curve, we observe that more than 95\% of the testing images are successfully predicted within the error range of 1.3 pixels, and over 98\% of the test set is well estimated within an error range of 1.4 pixels.

Fig. \ref{qualitative_results} presents a qualitative comparison between different approaches. From the figure, we observe that the predicted landmarks and heatmaps from the proposed methods generally align very well with the ground truth, and are less prone to ambiguities when landmarks are in close proximity (for instance, the upper lip).


\subsection{Ablation Studies}
\label{Ablation Studies}
\subsubsection{Impact of Different Regularization Constraints}
\label{Impact of Different Regularization Constraints}
In this section, we present an ablation study to analyze the effectiveness of each proposed regularization constraint, in which different combined regularization constraints are also added for evaluation. In addition to the original  four regularization constraints, we study the effectiveness of combinations of both the classifier and encoder outputs (\emph{RTL-CO-EO}), the classifier output with spatial attention map (\emph{RTL-CO-SAM}), and the classifier output with channel attention map (\emph{RTL-CO-CAM}).

\begin{figure}[t]
	\centering
	\includegraphics[clip, trim=0.4cm 0.01cm 1.8cm 2.7cm,width=0.85\linewidth]{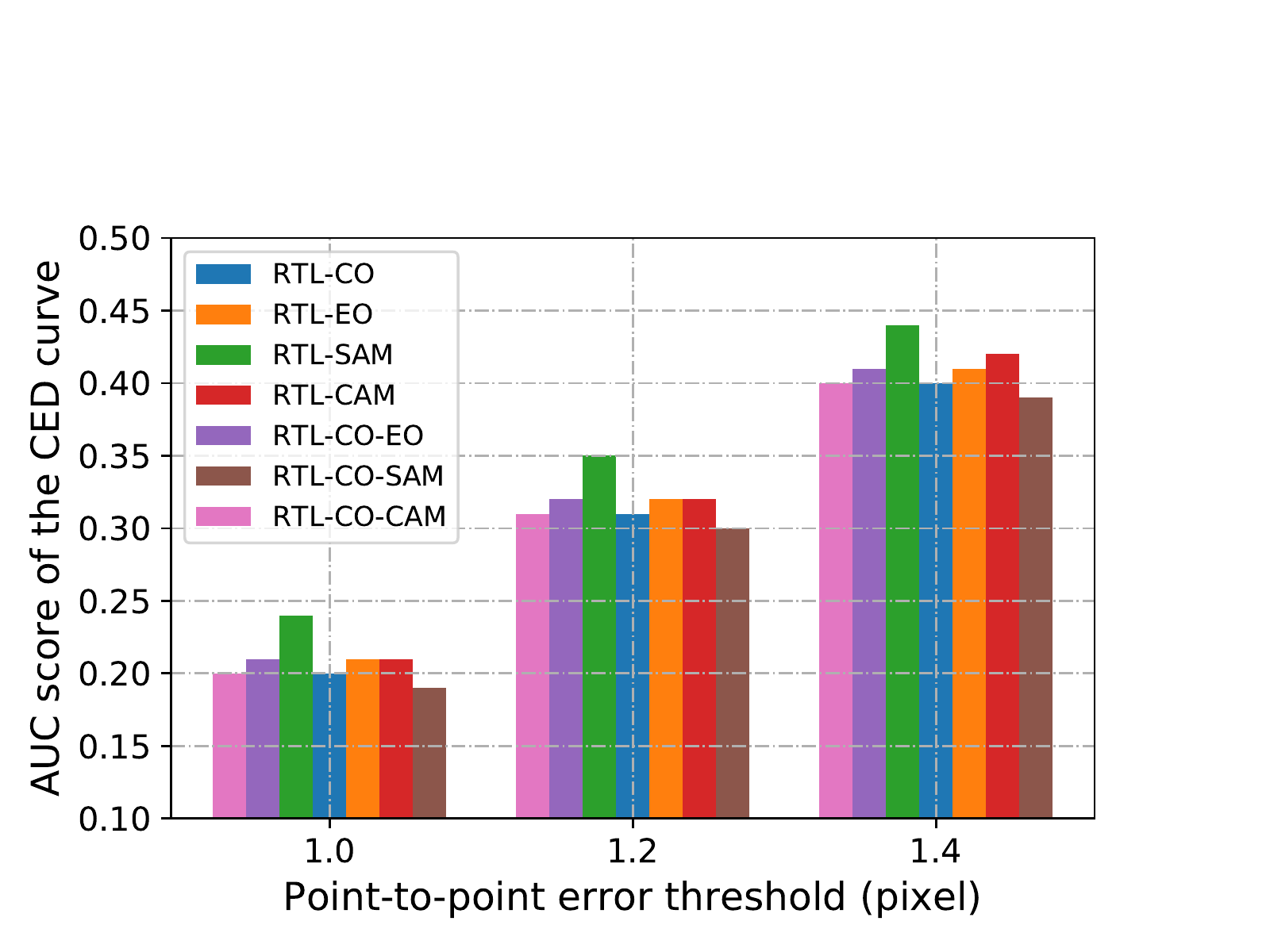}
	\caption{AUC scores of cumulative error distribution (CED) curve versus point-to-point error threshold for different proposed regularization constraints. Best viewed in color version.}
	\label{failure-rate}
\end{figure}%

Table \ref{table2} presents a detailed comparison of the different proposed approaches by listing the performance scores per individual landmark. From the results, we can see that \emph{RTL-EO} and \emph{RTL-SAM}  achieve the highest number of best scores $4$ and $5$ landmarks respectively. This indicates that preserving and highlighting the feature outputs with spatial information from intermediate layers can be potentially more beneficial for the localization task.
It can be also seen that for all the proposed variants the MSE ranged from 0.59 pixels to 1.58 pixels for different landmarks, where landmarks L1 and L2 corresponding to the exocanthion and 
endocanthion are well estimated with relatively lower scores than the others. 
However, similar to the visual findings in Fig. \ref{qualitative_results}, landmarks L7 and L9 marked as the crista philtri on both sides are seen as the most challenging cases.

Fig. \ref{failure-rate} compares the effect of modelling with the different proposed regularization constraints using the AUC score calculated from the CED curve.
Interestingly, the results indicate that combined approaches which introduce additional computational cost performed similarly or even worse than those of  individual ones. 
\begin{table}[t]
	\caption{Ablation studies on different pre-trained source tasks. A failure case here is considered if the point-to-point Euclidean error is greater than $1.2$ pixel. Best results are marked in bold.}
	\centering
	\label{table3}
	\resizebox{0.95\linewidth}{!}{
		\begin{tabular}{c|c|c|c}
			\toprule
			\multirow{2}{*}{Method}  & \multirow{2}{*}{MSE $\pm$ Std ($\downarrow$)}   & \multirow{2}{*}{FR ($\downarrow$)} & \multirow{2}{*}{AUC ($\uparrow$)}  \\
			&&  &   \\
			
			\hline
			FT-ImageNet \cite{simple-baseline}& 0.85$\pm$0.37& 9.68\% &   0.31   \\
FT-YoutubeFaces \cite{simple-baseline}& 	0.86$\pm$0.30& 	8.06\% &   	0.30  \\
			FT-VGGFace2 \cite{simple-baseline} & 0.86$\pm$0.24& 10.65\% &   0.29   \\
				\textit{RTL-SAM-ImageNet} & 0.82$\pm$0.32 & 8.39\%&  0.33  \\
				\textit{RTL-SAM-YoutubeFaces} & 	0.84$\pm$0.25 & 	8.06\% &   	0.31  \\
			\textit{RTL-SAM-VGGFace2}& \textbf{0.78}$\pm$\textbf{0.22} & \textbf{4.84\%} &  \textbf{0.35}  \\
			
			\bottomrule
		\end{tabular}
	}	
\end{table}
\subsubsection{Impact of Different Source Tasks}
\label{Impact of Different Pre-trained Source Tasks}
To investigate the impact of different source tasks on the proposed framework, we re-implement methods FT \cite{simple-baseline} and our proposed \textit{RTL-SAM} using the network ResNet-50 pre-trained on ImageNet \cite{ImageNet} and a sub-dataset of YoutubeFaces \cite{youtube-faces} (facial alignment with 3462 images). We denote the evaluated methods as FT-ImageNet, FT-YoutubeFace,  \textit{RTL-SAM-ImageNet} and  \textit{RTL-SAM-YoutubeFaces}. We then compare them with those previously obtained using  VGGFace2 \cite{VGGFACE2} pre-trained weights. We observe from Table \ref{table3}, that by utilizing  ImageNet and YoutubeFaces pre-trained weights, our proposed regularization strategy \textit{RTL-SAM} outperforms FT in most evaluated metrics, which further confirms the effectiveness of our proposed method. Moreover, it is noteworthy that the proposed method with VGGFace2 pre-trained weights (\textit{RTL-SAM-VGGFace2}) achieves the best performance among all compared methods. Interestingly, \textit{RTL-SAM-YoutubeFaces} using the source task which is closely relevant to the target task, did not show advantage over \textit{RTL-SAM-ImageNet} and \textit{RTL-SAM-VGGFace2}.

\begin{figure}[t]
	\centering
	\setlength{\belowcaptionskip}{-0.3cm}
	\setlength{\abovecaptionskip}{-0.2cm}
	\includegraphics[width=1\linewidth]{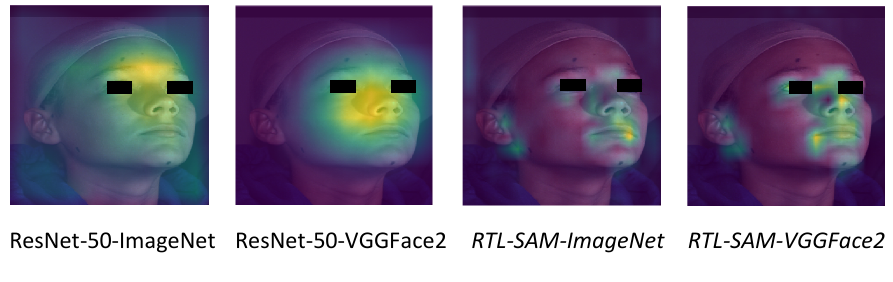}
	\caption{Illustration of top activation attention maps for models with different pre-trained source tasks: the source model (ResNet-50) pre-trained on ImageNet (ResNet-50-ImageNet), the source model (ResNet-50) pre-trained on VGGFace2 (ResNet-50-VGGFace2), the proposed regularization constraint on spatial attention maps transferred from ImageNet pre-trained weights (\textit{RTL-SAM-ImageNet}), and the proposed regularization constraint on spatial attention maps transferred from VGGFace2 pre-trained weights (\textit{RTL-SAM-VGGFace2}).}
	\label{impact-of-imagenet}
\end{figure}

To provide further insight on this finding, we visualize the top activation attention maps for the models derived with the source tasks of ImageNet and VGGFace2, as shown in Fig. \ref{impact-of-imagenet}.
We can see that the attention map generated from the source model (ResNet-50 \cite{resnet}) pre-trained on VGGFace2 \cite{VGGFACE2} (ResNet-50-VGGFace2) concentrates more on the desired facial region compared to ResNet-50-ImageNet.
After refining the pre-trained models to the target task, the attention map produced from \textit{RTL-SAM-VGGFace2} seems to be more informative than that of  \textit{RTL-SAM-ImageNet}  for inferring the anatomical landmarks. 

\subsubsection{Impact of $\lambda$}
\label{Impact of Weighting Parameter}
\begin{figure}[t]
	\centering
	\includegraphics[clip, trim=0.45cm 0.01cm 1.8cm 2.8cm,width=0.8\linewidth]{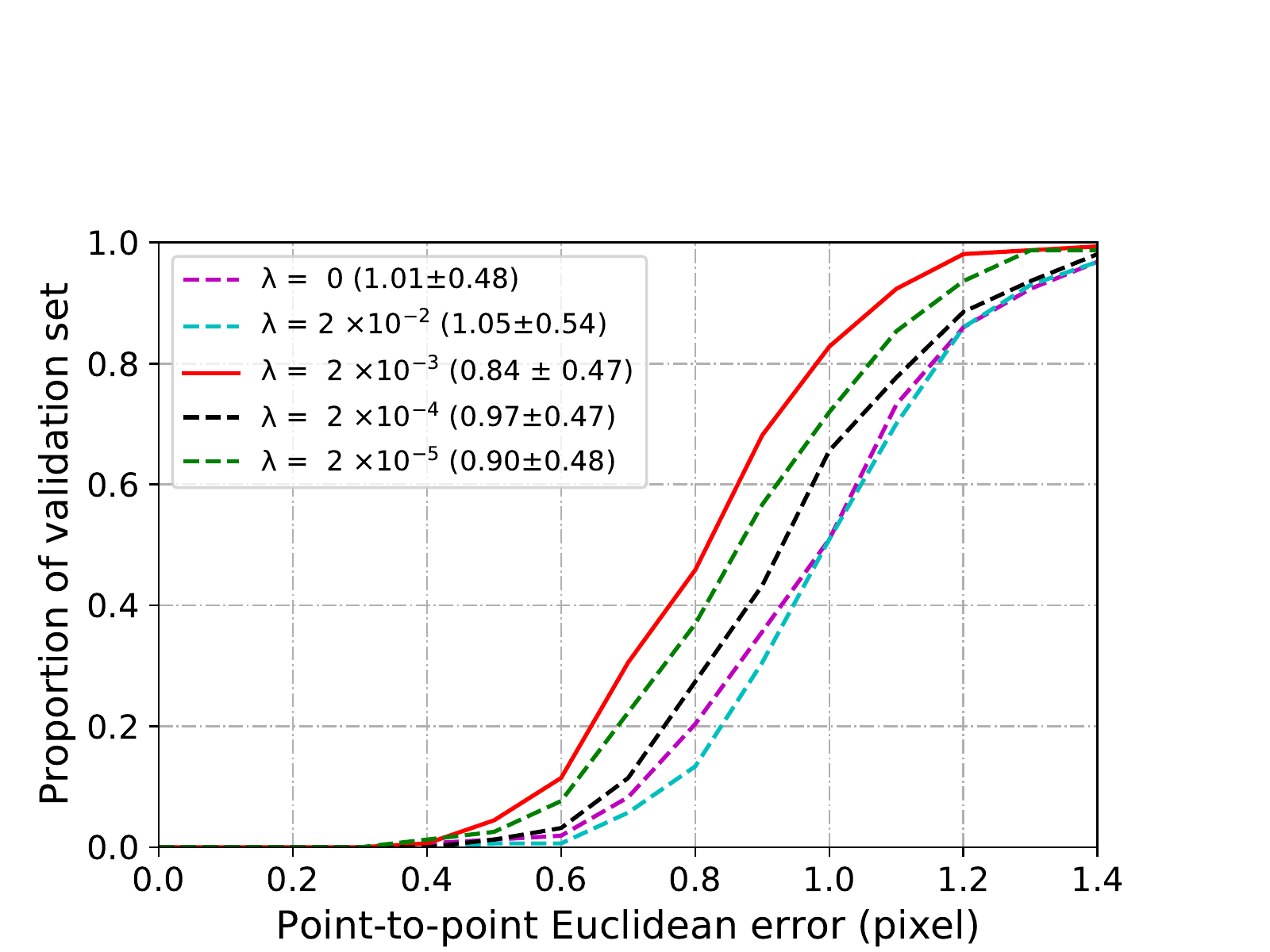}
	\caption{Illustration of CED curve on validation set for different values of weighting parameter $\lambda$, where corresponding results using MSE$\pm$Std are shown in the parenthesis.}
	\label{parameter_analysis}
\end{figure}%

To study the influence of weighting parameter choice on transfer learning regularization, we test a set of values for the parameter  $\lambda$ ranging from $0$ to $2\times 10^{-5}$ on the validation set using the proposed \emph{RTL-EO}, which has the simplest implementation among proposed methods. Note that when $\lambda = 0$, the designed model is evaluated without regularization, which is equivalent to standard fine tuning.
Fig. \ref{parameter_analysis} presents the results for varying parameter $\lambda$.  This shows that the proposed approach reduces the localization error when adjusting the value of $\lambda$ from $2\times 10^{-5}$ to  $2\times 10^{-3}$.
We also observe that the setting of $\lambda = 2\times 10 ^{-3}$, denoted as the red solid line, achieves the best performance considering both the CED curve and MSE.
However, a larger value of $\lambda = 2\times 10^{-2}$, regarded as a strong regularization during training, could affect model convergence 
for the landmark detection. 
\subsubsection{Impact of $\sigma$}
\label{Impact of  parameter sigma}
\begin{figure}[t]
	\centering
	\includegraphics[clip, trim=0.45cm 0.01cm 1.8cm 2.8cm,width=0.8\linewidth]{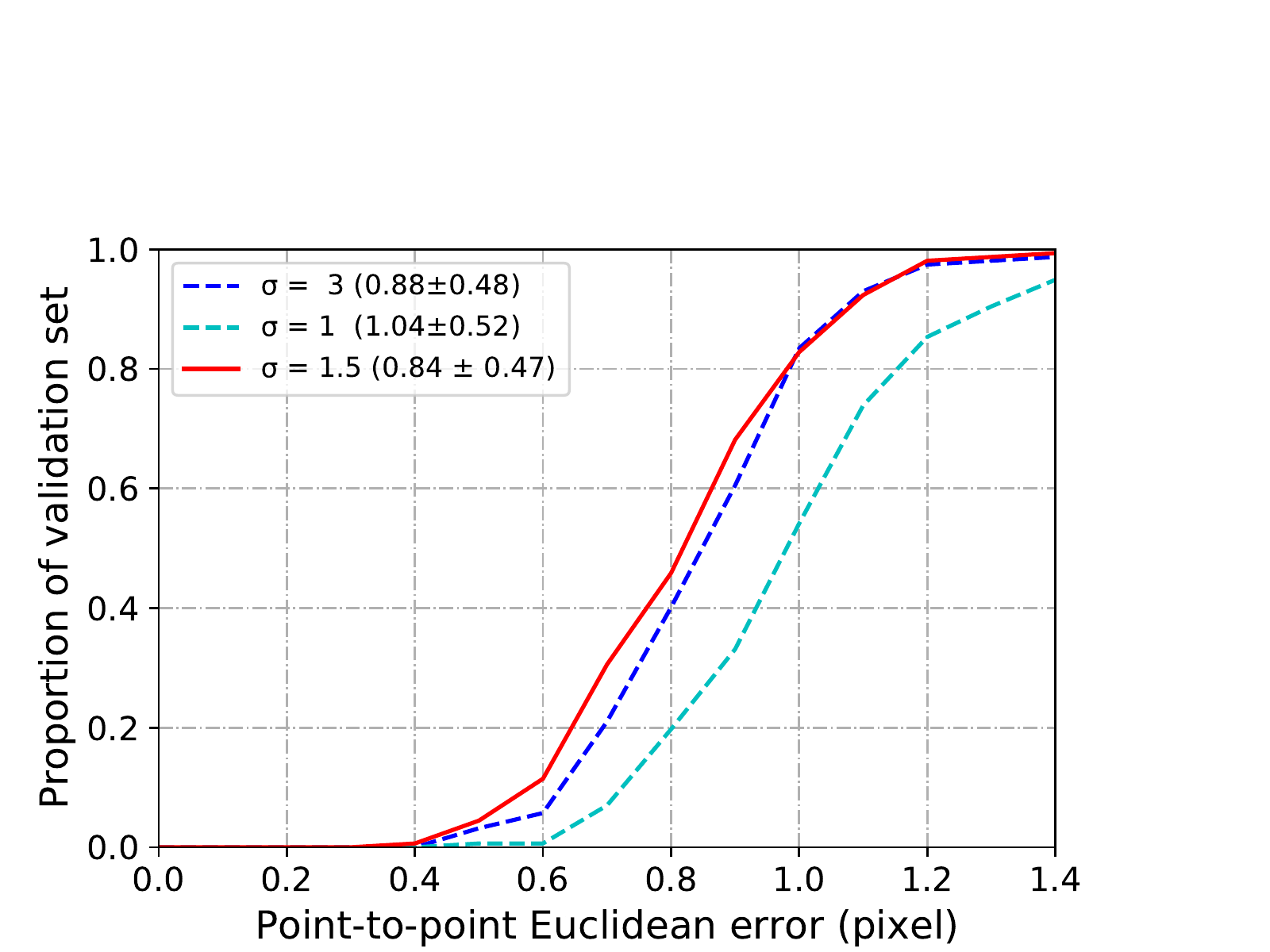}
	\caption{Illustration of CED curve on validation set for different values of kernel scaling parameter $\sigma$, where corresponding results using MSE$\pm$Std are shown in the parenthesis.}
	\label{sigma_analysis}
\end{figure}%
Here we conduct a sensitivity analysis to study the impact of Gaussian kernel scaling parameter $\sigma$ on the keypoint localization accuracy. Similar to the above analysis, we also utilize the proposed \emph{RTL-EO} with $\lambda = 2\times 10 ^{-3}$ to test a range of values for the parameter  $\sigma$ on the validation set. 
As seen from Fig. \ref{sigma_analysis}, we observe that the best performance evaluated in both metrics of CED and MSE is obtained with $\sigma = 1.5$, and the results in general are varied within a reasonable range. 

\section{Discussion}
\label{Discussion}
\subsection{Strengths and Limitations}
In this work, we have shown the importance of transferring knowledge from a pre-trained source classification model for learning the anatomical landmark detection with a limited availability of training data. This observation was validated by the experimental study in Section \ref{Comparison with Existing Approaches}, where transfer learning methods \cite{simple-baseline}, achieve noticeably better performance than other compared solutions. 
More importantly, the strength of this study is that the proposed methods are shown to better preserve facial semantics embedded in the source model for learning the anatomical landmark detection compared to finetuning. 
This can be best seen from the evaluations both quantitatively and qualitatively in Section \ref{Comparison with Existing Approaches}, where our proposed methods outperform the method of FT \cite{simple-baseline}. 
This property of  the proposed framework is particularly useful to reduce ambiguity in landmark detection, where landmarks which are in close proximity can be semantically distinguished by reusing pre-computed facial semantics. 

However, the prerequisite for the proposed framework is the utilization of a pre-trained  source task model, which may not be always available for other clinical applications.  
In Section \ref{Impact of Different Pre-trained Source Tasks}, we have demonstrated that the proposed approach performs better when incorporating the knowledge from VGGFace2 which is more relevant to the target task compared to that of ImageNet. 
Another important finding is that the proposed knowledge transfer with larger source datasets (ImageNet 14 million images and VGGFace2 3.31 million images) can generalize better on the target task regardless of task relevance. 

Moreover, evaluated results in Section \ref{Impact of Weighting Parameter} have shown that the proposed  framework with different settings of the weighting parameter has a generally positive impact on the transfer learning process for anatomical landmark detection, but these empirical configurations may be only specific to the addressed clinical application. Hence, how to adaptively determine the required amount of domain knowledge for a given target task remains an open question.  

\subsection{Translational Potential} 
The motivating clinical need for accurate, anatomical landmark placement in fetal alcohol syndrome (FAS) analysis is 1) to identify linear and angular facial anthropometric measurements such as palpebral fissure length (PFL)
and 2) to provide a surface-based analysis of facial form to compare an individual to unaffected controls.
We link this need to our research findings. The proposed framework delivered particularly good performance for landmarks L1 and L2 (exocanthion and 
endocanthion), as presented in Section \ref{Impact of Different Regularization Constraints}, which is important for accurate estimation of PFL. 
Previous analysis using a dense surface model (DSM) for the surface-based study of facial shape has shown the ability to identify subtle facial dysmorphism across FAS \cite{FAS, three-facial,FAS-curvature,Combined-FAS}. However, to date, their construction has been reliant upon the manual placement of a sparse set of anatomical landmarks. 
Based on the evaluations in Fig. \ref{ced-results},  more than 95\% of the testing cases have been accurately estimated by the proposed framework, where \emph{RTL-SAM} and \emph{RTL-EO} perform the best.  
We believe that the proposed framework can provide an automated alternative to the manual anthropometric placement, and can be readily embedded in a clinical prediction tool owing to its real-time speed about 44 frames per second (FPS). Specifically the proposed methods of \emph{RTL-SAM} and \emph{RTL-EO} are recommended  for the integration of a DSM into the clinical workflow for better identification of subjects with FAS. 

\section{Conclusions}
\label{Conclusions}
In this paper, we presented a new regularized transfer learning framework to address the problem of anatomical landmark detection in FAS.
The proposed learning approach reused the rich visual semantics from a domain-similar source model as guidance for learning the landmark detector. Four regularization constraints were developed and investigated in the proposed framework, including constraining the feature outputs from classification and intermediate layers, as well as matching activation attention maps in both the spatial and channel levels.
Experimental results show that the proposed learning framework performs well with limited training samples and outperforms other compared solutions.
The generalization of the proposed learning framework to other clinical applications with the availability of domain or medical expert knowledge would be an interesting research avenue. Future work will integrate the automated anatomical landmarks with a dense surface model (DSM) into the clinical workflow for better early identification of those with FAS.

\bibliographystyle{IEEEtran}
\bibliography{journalbib}
\section{Appendix}
\label{Appendix}
\subsection{Visualization of Training Details}
\label{Visualization of Training Details}
\subsection{Analysis of Adding Skip Connections}
\label{Analysis of Adding Skip Connections}
The baseline network architecture \cite{simple-baseline} used in the proposed learning framework does not include skip connections like U-Net\cite{Unet}. Here, we run a pilot test to study whether adding the skip connections between the encoder and decoder can improve the keypoint localization accuracy. For simplicity, we modify one of the baseline methods FT-ImageNet for this analysis. The results in Table \ref{analysis-skip} show that both MSE$\pm$Std and AUC score of FT-ImageNet are further improved by adding the skip connections. This validates the effectiveness of inclusion skip connections and suggests that such network design should be preferred for high-resolution keypoint localization in the future.

\begin{table}[H]
	\caption{Analysis of adding skip connections. A failure case here is considered if the point-to-point Euclidean error is greater than $1.2$ pixel. Best results are marked in bold.}
	\centering
	\label{analysis-skip}
	\resizebox{0.9\linewidth}{!}{
		\begin{tabular}{c|c|c|c}
			\hline
			\multirow{2}{*}{Method} & \multirow{2}{*}{MSE $\pm$ Std ($\downarrow$)}   & \multirow{2}{*}{FR ($\downarrow$)} & \multirow{2}{*}{AUC ($\uparrow$)}  \\
			&&  &   \\
			
			\hline
			Without skip connections & 0.85$\pm$0.37& \textbf{9.68\%} &  0.31   \\
			With skip connections& \textbf{0.83}$\pm$\textbf{0.27}& 11.9\% &   \textbf{0.33}\\
			\hline
		\end{tabular}
	}	
\end{table}

\end{document}